\documentclass[10pt,twocolumn,letterpaper]{article}

\usepackage{iccv}
\usepackage{times}
\usepackage{epsfig}
\usepackage{graphicx}
\usepackage{amsmath}
\usepackage{amssymb}
\usepackage{booktabs}
\usepackage{url}
\usepackage{subfigure}
\usepackage{color}
\usepackage{multirow}

\usepackage[T1]{fontenc}
\usepackage[utf8]{inputenc}
\usepackage{authblk}

\usepackage[pagebackref=true,breaklinks=true,letterpaper=true,hyperfootnotes=true,colorlinks,bookmarks=false]{hyperref}

\iccvfinalcopy 

\def\iccvPaperID{3060} 

\ificcvfinal\fi

\begin{document}

\title{A2J: Anchor-to-Joint Regression Network for 3D {Articulated} Pose Estimation from a Single Depth Image}

\author{\vspace{-3.5 mm} Fu Xiong$^1$$^*$, Boshen Zhang$^1$$^*$, Yang Xiao$^1$$^\dag$, Zhiguo Cao$^1$, Taidong Yu$^1$, Joey Tianyi~Zhou$^2$, and Junsong Yuan$^3$ \\
 \vspace{-3.5 mm} $^1$  National Key Laboratory of Science and Technology on Multi-Spectral Information Processing, School of Artificial Intelligence and Automation, Huazhong University of Science and Technology\\
$^2$ IHPC, A*STAR, Singapore~~~~~~~$^3$CSE Department, State University of New York at Buffalo\\
\tt\small{xiongfu, zhangbs, Yang$\_$Xiao, zgcao, taidongyu}@hust.edu.cn,\\ \tt\small joey.tianyi.zhou@gmail.com, jsyuan@buffalo.edu
}

\maketitle
\ificcvfinal\thispagestyle{empty}\fi

\let\thefootnote\relax\footnotetext{*Fu Xiong and Boshen Zhang devote the equal contribution.}
\let\thefootnote\relax\footnotetext{\dag Yang Xiao is the corresponding author (Yang$\_$Xiao@hust.edu.cn).}

\begin{abstract}
For 3D hand and {body} pose estimation task {in} depth image, a novel anchor-based approach termed Anchor-to-Joint regression network (A2J) with the end-to-end learning ability is proposed. Within A2J, anchor points able to capture global-local spatial context information are densely set on depth image as local regressors for the joints. They contribute to predict the positions of the joints in ensemble way to enhance generalization ability. The proposed 3D {articulated} pose estimation paradigm is different from the state-of-the-art encoder-decoder based FCN, 3D CNN and point-set based manners. To {discover informative anchor points towards certain joint, anchor proposal procedure is also proposed for A2J.} Meanwhile 2D CNN (i.e., ResNet-50) is used as backbone network to drive A2J, without using time-consuming 3D convolutional or deconvolutional layers. The experiments on 3 hand datasets and 2 body datasets verify A2J's superiority. Meanwhile, A2J is of high running speed around 100 FPS on single NVIDIA 1080Ti GPU.
\end{abstract}



\vspace{-1.5 mm}\section{Introduction}

\begin{figure}
\centering
\includegraphics[height=6.35cm]{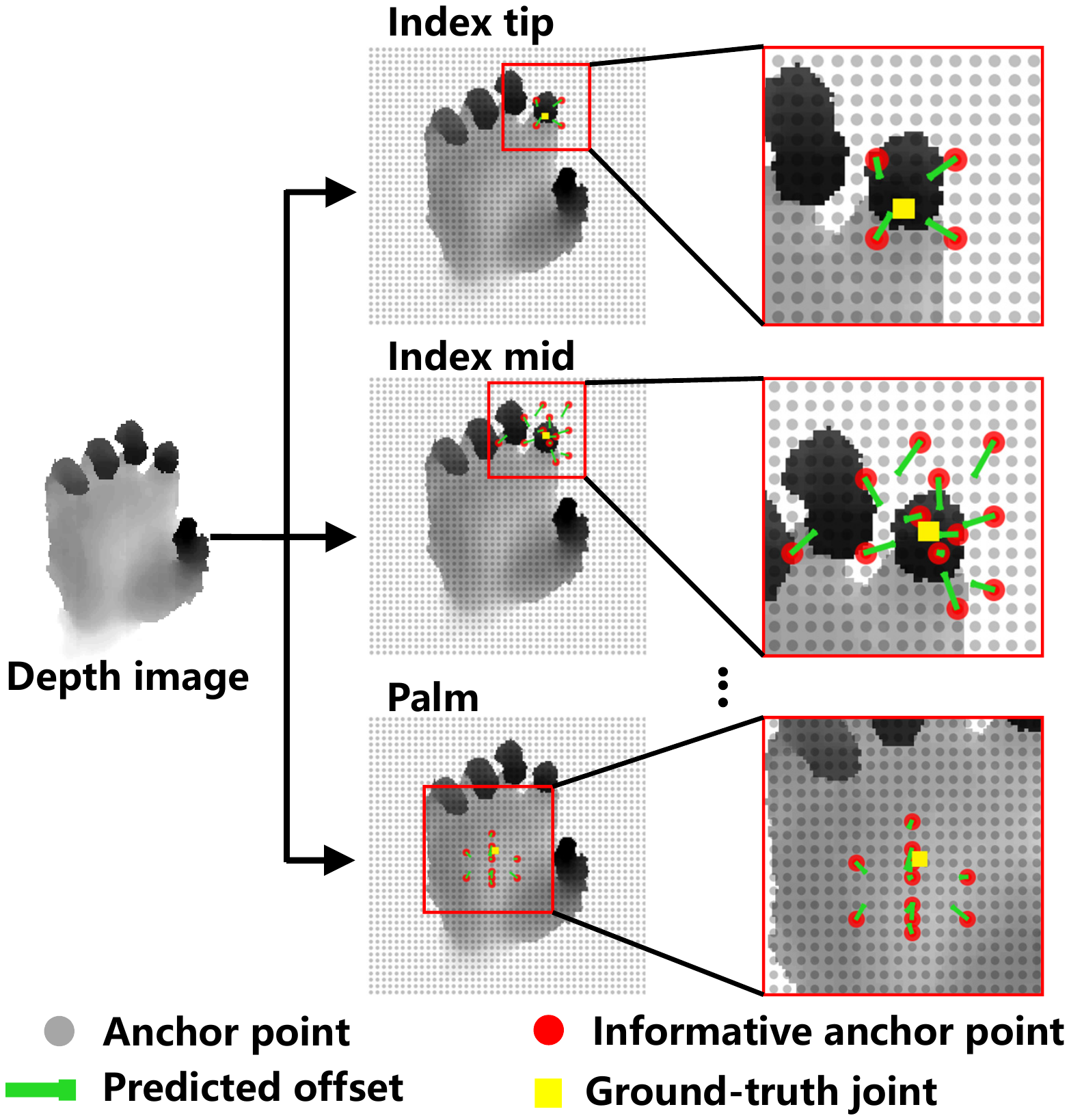}
\caption{The main idea of anchor-based 3D pose estimation paradigm within A2J. The densely set anchor points predict the positions of joints with weighted aggregation. The different joints possess the different informative anchor points of high weights (i.e., $>0.02$), which reveals A2J's adaptive characteristics.}
\label{fig:fig1}
\vspace{-0.0cm}
\end{figure}

With the emergence of low-cost depth camera, 3D hand and {body} pose estimation from a single depth image draws much attention from computer vision community with wide-range application scenarios (e.g., HCI and AR)~\cite{application, RF}. {Despite} recent remarkable progress~\cite{VI, K2HPD, V2V, REN966,REN466,Pose-REN,RF,RTW,CMB,CaiPose2019}, it is still a challenging task due to the issues of dramatic pose variation, high similarity among the different joints, self-occlusion, etc~\cite{VI,K2HPD,NYU}.

Most of state-of-the-art 3D hand and {body} pose estimation approaches rely on deep learning technology. Nevertheless, they still suffer from some defects. First, encoder-decoder based FCN manners~\cite{PHR,CPM,StackedHG,Openpose,K2HPD,CMB,V2V} are generally trained {with} non-adaptive ground-truth Gaussian heatmap for different joints and with relatively high computational burden. Meanwhile, {most of them cannot be fully end-to-end trained towards 3D pose estimation task~\cite{IntegralLoss}}. Secondly, 3D CNN models~\cite{3DCNN,Hand3D,V2V} are {difficult to train with costly voxelizing procedure}, due to the large number of convolutional parameters. Additionally, point-set based approaches~\cite{HandPointNet,P2P} require some extra time-consuming preprocessing treatments (e.g., point sampling).

Thus, we attempt to address 3D hand and {body} pose estimation problem {using a} novel anchor-based approach termed Anchor-to-Joint regression network (A2J). {The proposed A2J network has end-to-end learning ability}. The key idea of A2J is to \emph{predict {3D joint position} by aggregating the {estimation} results of multiple anchor points, in spirit of {ensemble learning to enhance} generalization {ability}}. Specifically, the anchor points can be regarded as the local regressors towards the joints from different viewpoints and {distances}. They are densely set on depth image to capture the global-local spatial context information together. {Each of them will contribute to regress the positions of all the joints, but with different weights. The joint is localized by aggregating the outputs of all the anchor points.} Since different joints may share the same anchor points, the articulated characteristics among them can be well maintained.

{For a specific joint,} not all of the anchor points contribute equally. Accordingly, an anchor proposal procedure is proposed to discover the informative anchor points towards the certain joint by weight assignment. During {training, both factors of estimation error of anchor points and spatial layout of informative anchor points are concerned}. In particular, \emph{the picked up informative anchor points are encouraged to uniformly surround the corresponding joint to alleviate overfitting}. Accordingly, the main idea of the proposed anchor-based 3D pose estimation paradigm within A2J is shown in Fig.~\ref{fig:fig1}. We can see that, generally different joints possess different informative anchor points. Furthermore, the visible ``index tip" joint holds few informative anchor points. While, the invisible ``index mid" joint and the ``palm" joint on the relatively flat area possess much more ones, {in order} to capture richer spatial contexts. This actually reveals A2J's adaptive property.

Technically, A2J network consists of 3 branches driven by 2D CNN backbone network (i.e., ResNet-50~\cite{ResNet}) without deconvolutional layers. In particular, the 3 branches take charges of predicting in-plain offsets between the anchor points and joints, estimating depth value of the joints, and informative anchor point proposal respectively. The main reasons {to} build A2J on 2D CNN for 3D pose estimation lie in 3 folders: (1) 3D information is already involved in depth image, using 2D CNN can still reveal 3D characteristics {of the original depth image data}; (2) compared to 3D CNN and point-set network, 2D CNN can be pre-trained on large-scale datasets (e.g., ImageNet~\cite{ImageNet}), which may help to enhance its visual pattern capturing capacity for depth image; (3) 2D CNN is of high running efficiency without time-consuming 3D convolution operation and preprocessing procedures (e.g., voxelizing and point sampling).


A2J is {experimented} on 3 hand datasets (i.e., HANDS 2017~\cite{Hands2017}, NYU~\cite{NYU}, and ICVL~\cite{ICVL}) and 2 {body} pose datasets (i.e., ITOP~\cite{VI} and K2HPD~\cite{K2HPD}) to verify its superiority. The experiments reveal that, both for 3D hand and {body} pose estimation tasks A2J generally outperforms the state-of-the-art methods on effectiveness and efficiency simultaneously. Meanwhile, A2J can online run with the high speed around 100 FPS on a single NVIDIA 1080Ti GPU.

The main contributions of this paper include:

$\bullet$ A2J: {an} anchor-based regression network for 3D hand and body estimation from a single depth image. It is of end-to-end learning capacity;

$\bullet$ An informative anchor proposal approach is proposed, concerning the joint position prediction error and anchor spatial layout simultaneously;

$\bullet$ 2D CNN without deconvolutional layers is used to drive A2J to ensure high running efficiency.

{A2J's code is available at~\url{https://github.com/zhangboshen/A2J}.}



\begin{figure*}[t]
\centering
\includegraphics[width=12cm]{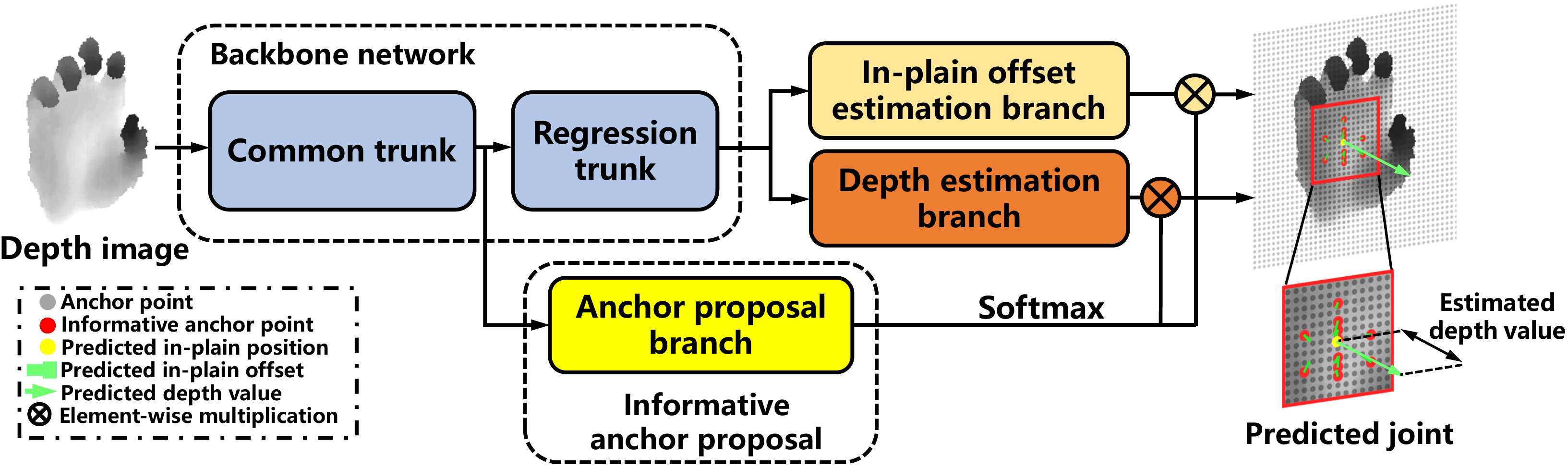}
\caption{The main technical pipeline of A2J. A2J consists of backbone network and 3 functional branches. The backbone network is built on ResNet-50. And, the 3 branches are in-plain offset prediction branch, depth estimation branch, and anchor proposal branch. }
\label{fig:A2Jpipline}
\end{figure*}

\section{Related Works} \label{Related Works}
The existing 3D hand and {body} pose estimation approaches can be mainly categorized into non-deep learning and deep learning based groups. The state-of-the-art non-deep learning based ones~\cite{RF,graphical_depth1,EVAL,databaseLookups} generally follow the 2-step technical pipeline of first extracting hand-crafted feature, and then executing classification or regression. One main drawback is that, hand-crafted feature is often not representative enough. This tends to lead non-deep learning based method to be inferior to deep learning based manner. Since the proposed A2J falls into deep learning group, next we will introduce and discuss this paradigm from the perspectives of 2D and 3D deep learning respectively.

\textbf{2D deep learning based approach.} Due to end-to-end working manner, deep learning technology holds strong fitting ability for visual pattern characterization. 2D CNN has already achieved great success for 2D pose estimation~\cite{Deeppose,Openpose,StackedHG,CPM,MSbaseline}. Recently it has also been introduced to 3D domain, resorting to global regression~\cite{REN966,REN466,DeepPrior,Deepprior++,Pose-REN,MultiView,VI} or local detection~\cite{NYU,LocalVolumetric,K2HPD,CMB,DenseReg} ways. The global regression manner cannot well maintain local spatial context information due to the global feature aggregation operation within fully-connected layers. Local detection based paradigm of promising performance generally chooses to address this problem via encoder-decoder model (e.g., FCN), setting local heatmap for each joint. Nevertheless, heatmap setting is still not adaptive for the different joints. And, the deconvolution operation is time consuming. Furthermore, most of the encoder-decoder based methods cannot be fully end-to-end trained~\cite{MSbaseline}.

\textbf{3D deep learning based approach.}
To better reveal the 3D property within depth image for performance enhancement, one recent research trend is to resort to 3D deep learning. The paid efforts can be generally categorized into 3D CNN based and point-set based families. 3D CNN based methods~\cite{3DCNN,Hand3D,V2V} voxelizes the depth image into volumetric representation (e.g., occupancy grid models \cite{Voxnet}). 3D convolution or deconvolution operation is then executed to capture 3D visual characteristics. However, 3D CNN is relatively hard to tune due to the large number of convolutional parameters. Meanwhile, 3D voxelization operation also leads to high computational burden both on memory storage and running time. Another way for 3D deep learning is point-set network~\cite{PointNet,PointNet++}, transferring depth image into point cloud as input. Nevertheless some time-consuming procedures (e.g., point sampling and KNN search) are required~\cite{PointNet,PointNet++}, which weakens running efficiency.

Accordingly, A2J belongs to 2D deep learning based group. The dense anchor points capture the global-local spatial context information in ensemble way, without using computationally expensive deconvolutional layers. 2D CNN is used as the backbone network for high running efficiency, also aiming to transfer knowledge from RGB domain.

 \begin{table}
\scriptsize
\begin{center}
\begin{tabular}{cc}
\toprule
Symbol & Definition \\
\hline\hline
$A$ & Anchor point set. \\
$a$ & Anchor point $a\in A$. \\
$J$ & Joint set.\\
$j$ & Joint $j\in J$. \\
$K$ & Number of joints. \\
$S(a)$& In-plain position of anchor point $a$.\\
$P_j(a)$ & Response of anchor $a$ towards joint $j$. \\
$O_j(a)$ & Predicted in-plain offset towards joint $j$ from anchor point $a$.\\
$D_j(a)$ & Predicted depth value of joint $j$ by anchor point $a$. \\
\bottomrule
\end{tabular}
\end{center}
\caption{Symbol definition within A2J.}
\label{tab:symbol}
\end{table}

\section{A2J: Anchor-to-Joint Regression Network}

The main technical pipeline of A2J is shown in Fig.~\ref{fig:A2Jpipline}. And, the symbols within A2J are defined in Table~\ref{tab:symbol}. A2J consists of 2D backbone network (i.e., ResNet-50), and 3 functional branches: in-plain offset estimation branch, depth estimation branch, and anchor proposal branch. The 3 branches predict $O_j(a)$, $D_j(a)$, and $P_j(a)$ respectively.

\begin{figure}
\centering
\includegraphics[height=3.4cm]{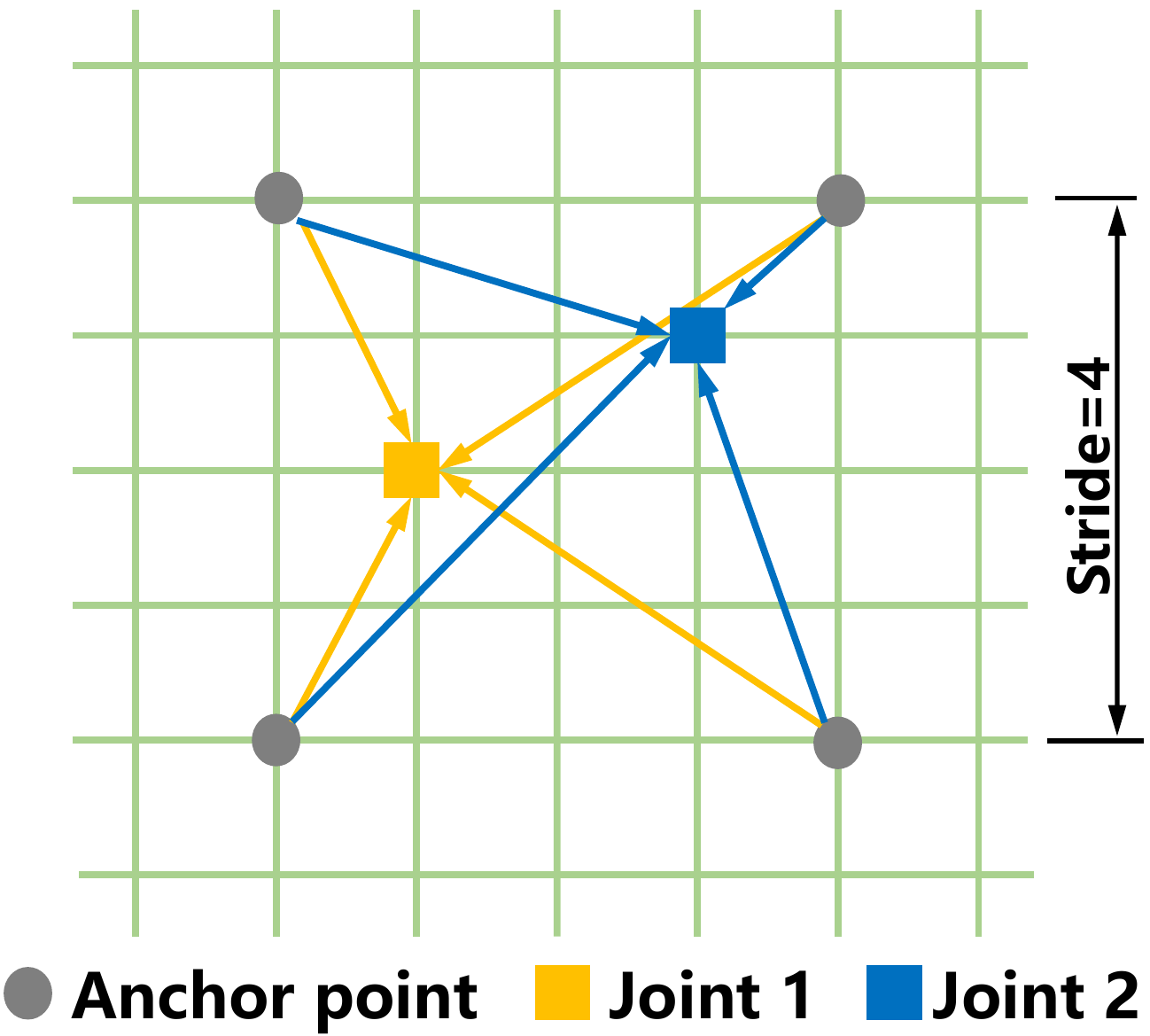}
\caption{The densely set anchor points on depth image. They will serve for predicting the positions of all joints in ensemble way.}
\label{fig:A2J}
\vspace{-0.2cm}
\end{figure}

Within A2J, anchor points are densely set up on the input depth images with stride $S_t=4$ pixels to capture the global-local spatial context information as in Fig.~\ref{fig:A2J}. Essentially, each of them serves as the local regressor to predict the 3D position of all the joints via in-plain offset prediction branch and depth estimation branch. For certain joint, it is finally localized by {aggregating} the outputs of all the anchor points. Concerning that maybe not all the anchor points contribute equally to certain joint, the anchor points will be assigned weights via anchor proposal branch to discover the informative ones. As consequence, the in-plain position and depth value of joint $j$ can be achieved as the weighted average of the outputs of all anchor points as:
\begin{equation}
\label{eq:A2J}
\begin{aligned}
\left\{ \begin{array}{l}
{{\hat S}_j} = \sum\limits_{a \in A} {{{\tilde P}_j}\left(a\right)\left({S\left(a\right)} + {O_j}\left(a\right)\right)} \\
{{\hat D}_j} = \sum\limits_{a \in A} {{{\tilde P}_j}(a){D_j}(a)}
\end{array} \right.,
\end{aligned}
\end{equation}
 where ${{\hat S}_j}$ and ${{\hat D}_j}$ indicate the estimated in-plain position and depth value of joint $j$; $\tilde P_j(a)$ can be regarded as the normalized weight of anchor point $a$ towards joint $j$ across all anchor points, and is acquired using softmax by:
 \begin{equation}
\label{eq:softmax}
\begin{aligned}
 \tilde P_j(a) = \frac{{{e^{{P_j}(a)}}}}{{\sum\limits_{a \in A} {{e^{{P_j}(a)}}} }}.
 \end{aligned}
\end{equation}
It is worthy noting that, the anchor point $a$ with $\tilde P_j(a)>0.02$ will be regarded as the informative anchor points for joint $j$. {The selected informative anchor points can reveal A2J's adaptive characteristics as in Fig.~\ref{fig:fig1}}. Joint position estimation loss and anchor point surrounding loss are used to supervise A2J's end-to-end training. Under their joint supervision, informative anchor points with the spatial layout that surrounds the joint will be picked up to enhance generalization ability. Next, we will illustrate the proposed A2J regression network and its learning procedure in details.

\subsection{A2J regression network}

Here, the 3 functional branches and backbone network within A2J will be illustrated in details respectively.

\subsubsection{In-plain offset and depth estimation branches} \label{sec:in-plain_depth_branch}
\begin{figure}
\centering
\includegraphics[height=4.8cm]{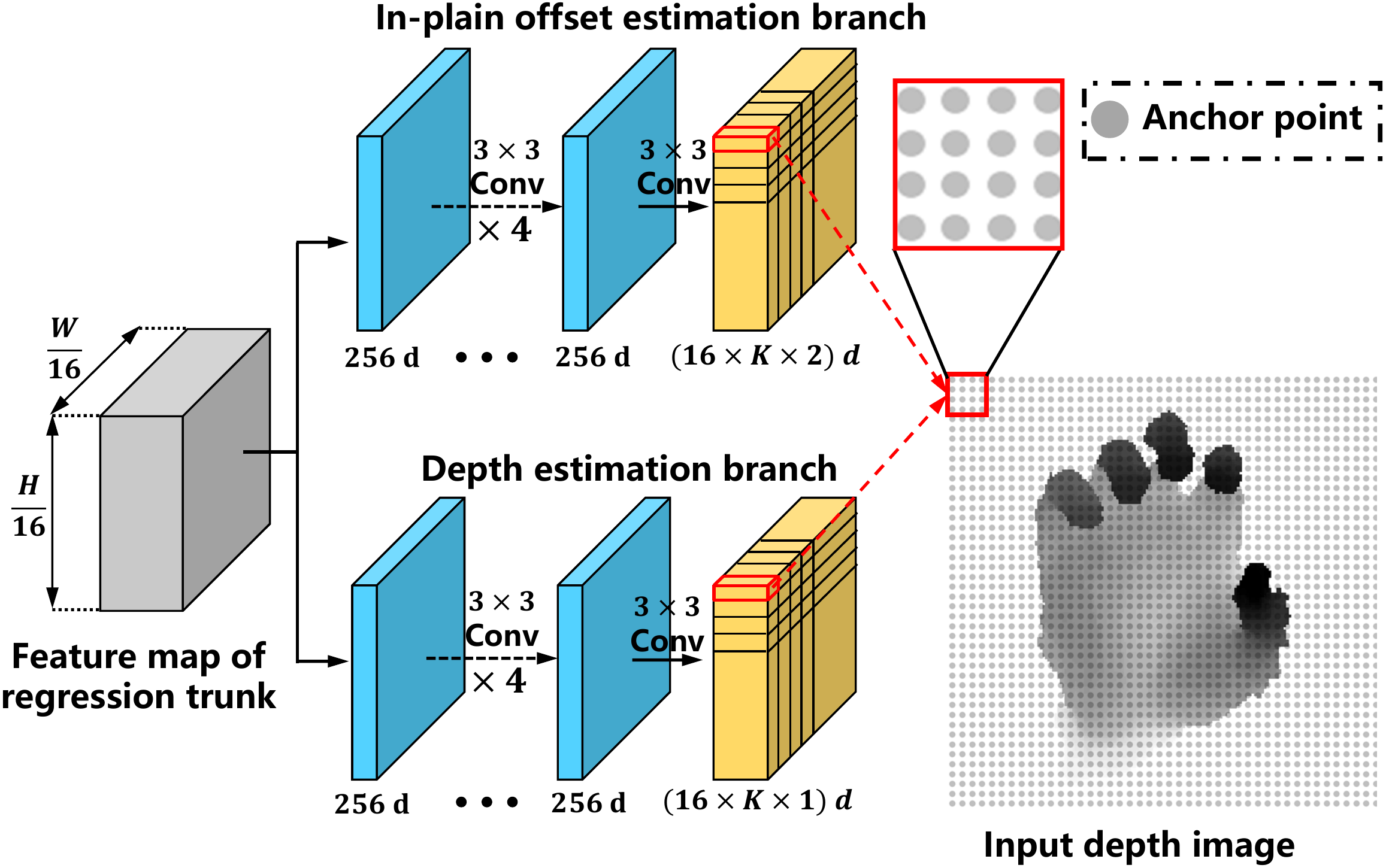}
\caption{In-plain offset and depth estimation branches. They both contain 4 intermediate convolutional layers with 256 channels, and 1 output convolutional layer (with $16\times K\times2$ or $16\times K\times1$ channels). $W$ and $H$ indicate width and height of the input depth image. $d$ means dimensionality.}
\label{fig:offsetbranch}
\end{figure}

Essentially, these 2 branches play the role of predicting the 3D positions of joints. {Since in-plain position estimation and depth estimation are of different properties, we choose to execute them separately.} Specifically, one is to estimate $O_j(a)$  between anchor points and joints. And, the other is to estimate $D_j(a)$ {towards joints}. As in Fig.~\ref{fig:offsetbranch}, they are built upon the output feature map of regression trunk within backbone network to involve semantic feature. Four $3\times3$ intermediate convolutional layers ({with BN and ReLU)} are consequently set to aggregate richer local context information without reducing in-plain size. Since the feature map is a $16\times$ downsampling of the input depth image on in-plain size (illustrated in Sec.~\ref{Backbone}) and anchor point setting stride $S_t=4$ as in Fig.~\ref{fig:A2J}, one feature map point corresponds to $4\times4=16$ anchor points on depth image. An output convolutional layer with the feature map in-plain size is then set towards all the 16 corresponding anchor points in column-wise manner. Suppose $K$ joints exist, in-plain offset estimation branch is of $16 \times K\times2$ output channels. And, depth estimation branch is of $16\times K\times1$ output channels.

\subsubsection{Anchor proposal branch} 

\begin{figure}[t]
\centering
\includegraphics[height=4.8cm]{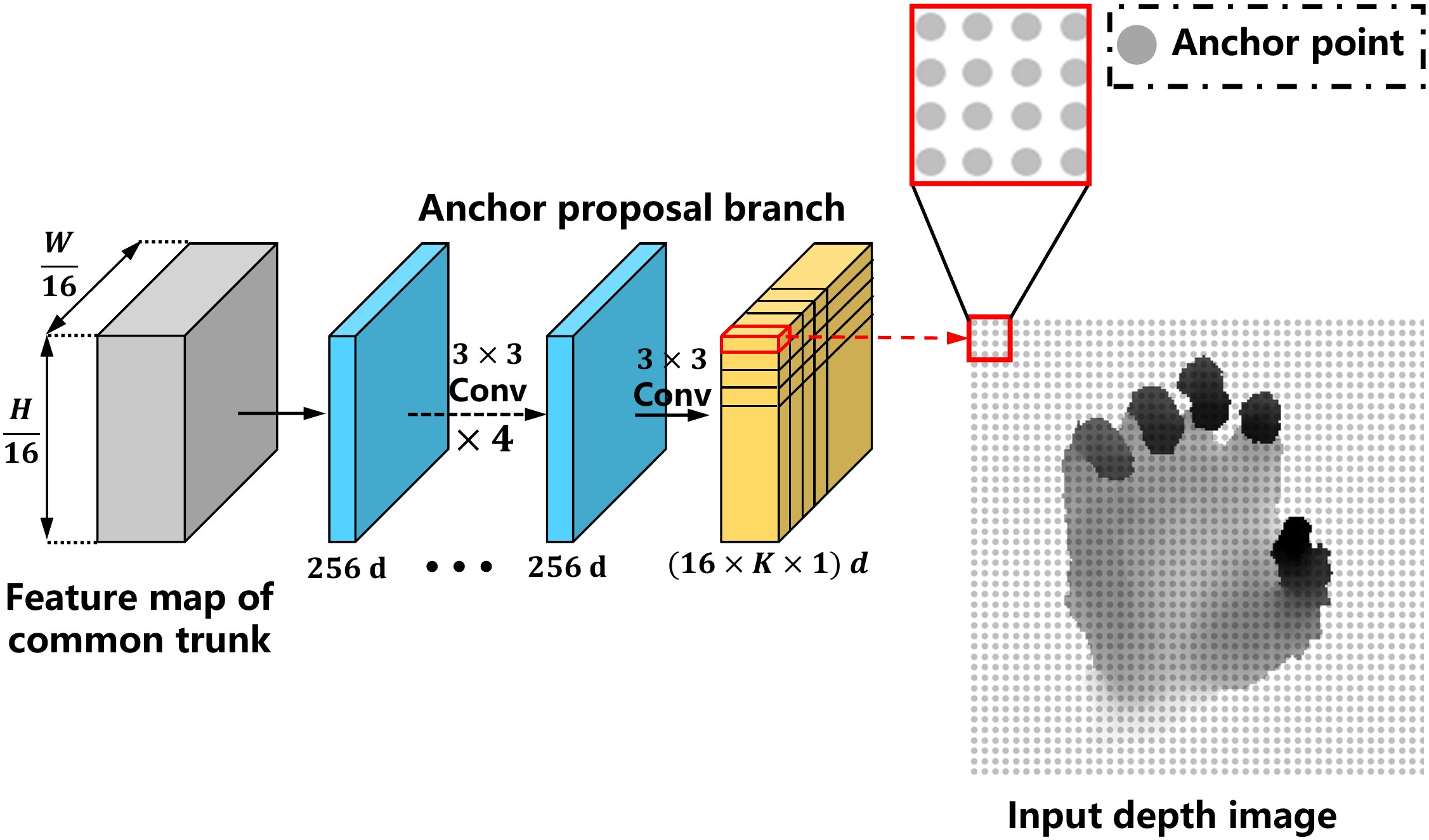}
\caption{Anchor proposal branch with 4 intermediate convolutional layers with 256 channels, and 1 output convolutional layer with $16 \times K \times1$ channels. $W$ and $H$ indicate width and height of the input depth image. $d$ means dimensionality.}
\label{fig:anchorbranch}
\end{figure}

This branch discovers informative anchor points for the certain joint by weight assignment as Eqn.~\ref{eq:softmax}. As in Fig.~\ref{fig:anchorbranch}, anchor proposal branch is built upon the output feature map of common trunk within backbone network to involve relatively fine feature. As the 2 branches introduced in Sec.~\ref{sec:in-plain_depth_branch}, 4 intermediate convolutional layers and 1 output convolutional layer are consequently set for predicting $P_j(a)$ for the anchor points  without losing in-plain size. Accordingly, the output layer of this branch is of $16\times K\times1$ channels.

\subsubsection{Backbone network architecture} \label{Backbone}

ResNet-50~\cite{ResNet} pre-trained on ImageNet is used as the backbone network. In particular, layers 0-3 correspond to the common trunk in Fig.~\ref{fig:A2Jpipline}. And, layer 4 corresponds to regression trunk. Some modifications are executed to make ResNet-50 more suitable for pose estimation. First, the {convolutional stride} in layer 4 is set to 1. Consequently, the output feature map of layer 4 is a 16$\times$ downsampling of the input depth image on in-plain size. Compared with the raw ResNet-50 with 32$\times$ downsampling, more fine spatial information can be maintained in this way. Meanwhile, the convolution operation within layer 4 is revised as the dilated convolution with a dilation of 2 to enlarge receptive field.

\subsection{Learning procedure of A2J}

To generate input of A2J, we follow~\cite{V2V} and use center points to crop the hand region from depth image. For {body} pose, we follow~\cite{RMPE} and use bounding box to crop the {body} region. For joint $j$, in-plain target $T^i_{j}$ denotes the 2D ground-truth in pixel coordinate transformed according to the cropped region. To make $T^i_{j}$ and depth target $T^d_{j}$ be in comparable magnitude, we transform the ground-truth depth $Z_j$ of joint $j$ as:
\begin{equation} \label{eq:depthGT}
\begin{aligned}
{T^d_{j} = \mu (Z_j - \theta)},
\end{aligned}
\end{equation}
where $\mu$ and $\theta$ are the transformation parameters. For hand pose $\mu$ is set to 1, and $\theta$ is set to the depth of center points. For {body pose} $\mu$ is set to 50 and $\theta$ is set as 0, since we do not have depth center. During test, the prediction result will be warpped back to world coordinate. A2J is then trained under the joint supervision of 2 loss functions: joint position estimation loss and informative anchor point surrounding loss. Next, we will illustrate these 2 loss functions in details.

\subsubsection{Joint position estimation loss}

Within A2J, the anchor points serve as the local regressors to predict the 3D position of joints in ensemble way. This objective loss can be formulated as:
\begin{equation}
\label{eq:loss1}
\begin{aligned}
loss_1 &= \alpha \sum\limits_{j \in J} {L_{\tau_1}(\sum\limits_{a \in A} {\tilde P_j(a) (S(a) + {O_j}(a)) - {T^i_{j}}})}  \\
&+ \sum\limits_{j \in J} {L_{\tau_2}(\sum\limits_{a \in A} {\tilde P_j(a) {{D_j}(a) - {T^d_{j}}}})},
\end{aligned}
\end{equation}
where $\alpha=0.5$ is the factor to balance in-plain offset and depth estimation task; $T^i_{j}$ and $T^d_{j}$ are the in-plain and depth targets position of joint $j$; and $L_{\tau}(\cdot)$ is the {${\rm smooth}_{L1}$} like loss function~\cite{FasterRCNN} given by:
\begin{equation}
\label{eq:smoothL1}
\begin{aligned}
{L_\tau(x) } = \left\{ \begin{array}{l}
\frac{1}{{2\tau }}{x^2},{\qquad \textnormal{for} \ }\left| x \right| < \tau ,\\
\left| x \right| - \frac{\tau }{2},{\quad }\textnormal{otherwise}.
\end{array} \right.
\end{aligned}
\end{equation}
In Eqn.~\ref{eq:loss1}, $\tau_1$ is set to 1 and $\tau_2$ is set to 3 since the depth value is relatively noisy.


\subsubsection{{Informative anchor point surrounding loss}}

\begin{figure}[t]
\centering
\subfigure[Hand cases from NYU dataset] {\label{sub_hand} \includegraphics[width=0.36\textwidth]{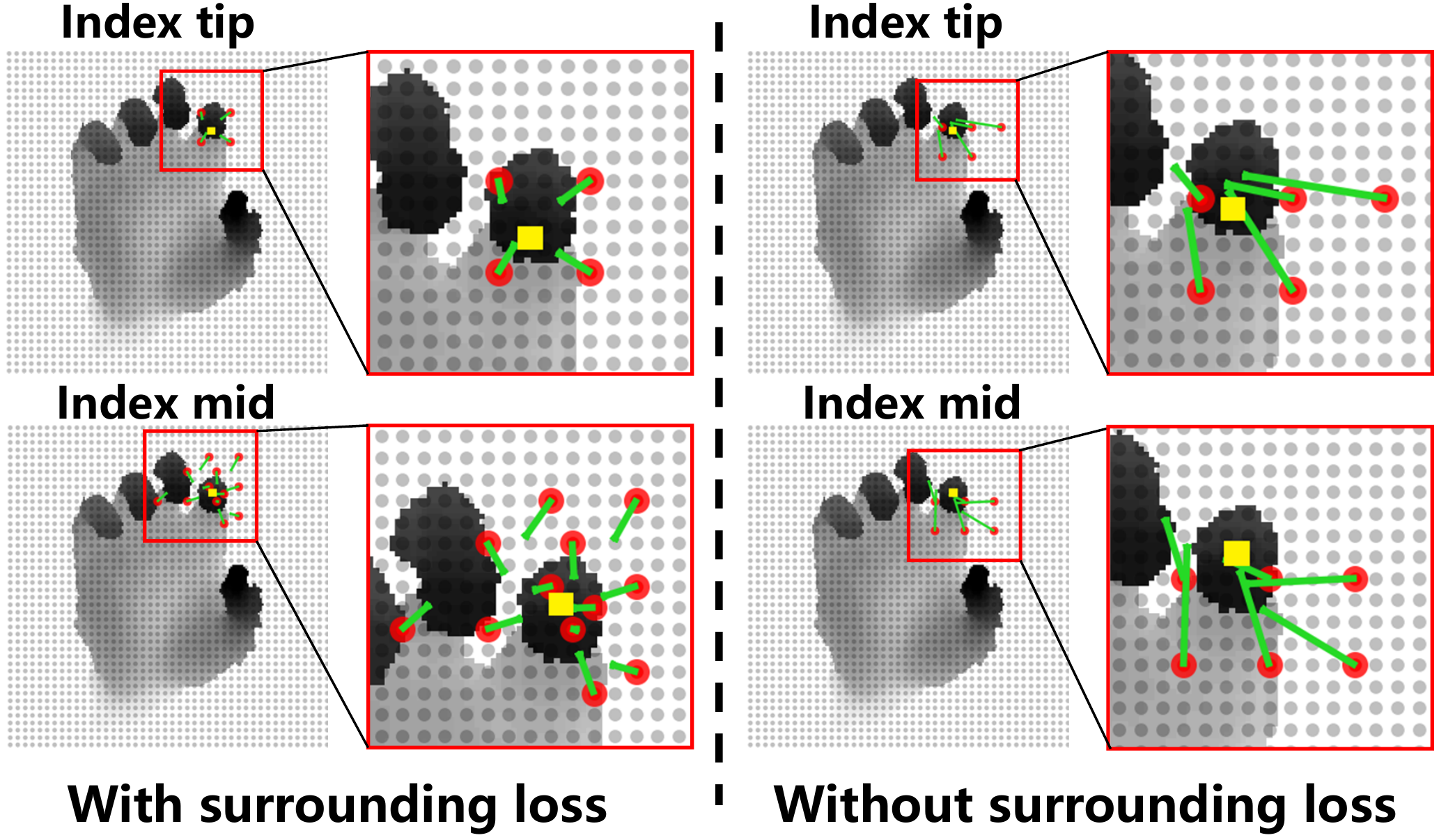}}
\subfigure[Body cases from ITOP front-view dataset] {\label{sub_human} \includegraphics[width=0.36\textwidth ]{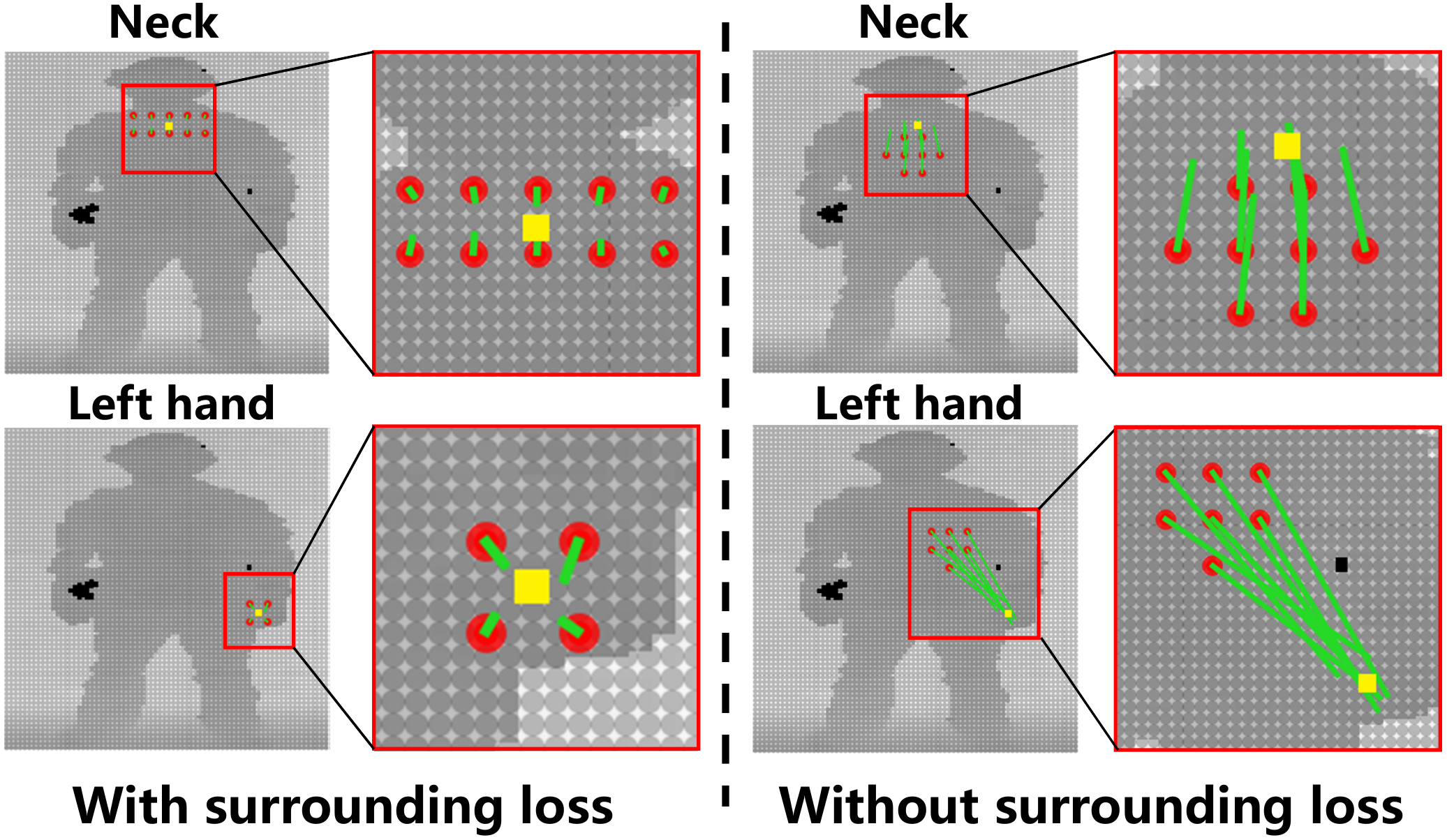}}

\caption{Effectiveness of anchor point surrounding loss. Grey dot denotes anchor point. Red dot indicate informative anchor point. Green arrow represent in-plain offset. Yellow square corresponds to ground-truth joint.}
\label{fig:surroundloss}
\end{figure}

To enhance the generalization ability of A2J, we intend to let the picked up informative anchor points locate around the joints, in spirit of observing the joints from multiple viewpoints simultaneously. Hence, the informative anchor point surrounding loss is defined by us as:
\begin{equation}
\label{eq:loss2}
\centering
\begin{aligned}
loss_2 = \sum\limits_{j \in J} {L_{\tau_1}( {\sum\limits_{a \in A} {\tilde P_j(a)S(a)}  - {T^i_{j}}} )}.
\end{aligned}
\end{equation}
To reveal its effectiveness, we show the informative anchor point spatial layouts with and without using it both for hand { and body pose cases in Fig.~\ref{fig:surroundloss}}. It can be seen that, informative anchor point surrounding loss can essentially help to alleviate viewpoint bias. Its quantitative effectiveness will also be verified in Sec.~\ref{ablation:component}.

\subsubsection{End-to-end training}

The 2 loss functions above jointly supervise the end-to-end learning procedure of A2J, which is formulated as:
\begin{equation}
\label{eq:loss}
\begin{aligned}
loss = \lambda loss_1 + loss_2,
\end{aligned}
\end{equation}
where $loss$ is the loss in all; and $\lambda=3$ is the weight factor to balance $loss_1$ and $loss_2$.

\section{Experiments} \label{Experiments}


\subsection{Experimental setting} \label{Experimente:1}

\subsubsection{Datasets} \label{Experimente:1.1}


\noindent \textbf{HANDS 2017 dataset}~\cite{Hands2017}. It contains 957K training and 295K test depth images sampled from BigHand 2.2M~\cite{BigHand} and First-Person Hand Action~\cite{Hands2017} datasets. The ground-truth is the 3D coordinates of 21 hand joints.

\noindent\textbf{NYU Hand Pose Dataset}~\cite{NYU}. It contains 72K training and 8.2K test depth images with 3D annotation on 36 hand joints. Following~\cite{Pose-REN,REN466,REN966,V2V}, we pick 14 of the 36 joints from frontal view for evaluation.

\noindent\textbf{ICVL Hand Pose Dataset}~\cite{ICVL}. It contains 22K training and 1.5k test depth images. It is augmented to 330K samples by in-plane rotations. 16 hand joints are annotated.

\noindent\textbf{ITOP { Body} Pose Dataset}~\cite{VI}. It contains 40K training and 10K test depth images both for the front-view and top-view tracks. Each depth image is labelled with 15 3D joint locations of human body.

\noindent\textbf{K2HPD { Body} Pose Dataset}~\cite{K2HPD}. It contains about 100K depth images. 19 human body joints are annotated with the in-plain manner.

\subsubsection{Evaluation metric} \label{Experimente:1.2}

For hand, the \textbf{average 3D distance error} {and \textbf{percentage of success frame metrics}~\cite{V2V}} are used as the evaluation criteria. For body,\textbf{ Percent of Detected Joints (PDJ)}~\cite{K2HPD, VI} and \textbf{mean average precision (mAP)} with 10-cm rule~\cite{K2HPD, VI} are used for evaluation.


\subsubsection{Implementation details} \label{Experimente:1.3}

A2J network is implemented using PyTorch. The input depth image is cropped and resized to a fixed resolution (i.e., $176\times176$ for hand, and $288\times288$ for body). Random in-plain rotation and random scaling for both in-plain and depth dimension are executed for data augment. Random Gaussian noise is also randomly added with the probability of 0.5 for data augment. We use Adam as the optimizer. The learning rate is set to 0.00035 with a weight decay of 0.0001 in all cases. A2J is trained on NYU for 34 epochs with a learning rate decay by 0.1 every 10 epoch, and for 17 epochs on ICVL and HANDS 2017 with a learning rate decay by 0.1 every 7 epoch. For 2 human body datasets, the epoch for training is set as 26 with a learning rate decay by 0.1 every 10 epoch.

\begin{table}
\scriptsize
\begin{center}
\begin{tabular}{ccccc}
\toprule
Methods & AVG & SEEN & UNSEEN & FPS\\
\hline\hline
Vanora \cite{RCN3D} & 11.91 & 9.55 & 13.89 & -\\
THU VCLab \cite{THUVCLab} & 11.70 & 9.15 & 13.83 & -\\
Oasis \cite{HandPointNet} & 11.30 & 8.86 & 13.33 & 48\\
RCN-3D \cite{RCN3D} & 9.97 & 7.55 & 12.00 & - \\
V2V$^*$ \cite{V2V} & 9.95 & 6.97 & 12.43 & 3.5 \\
\midrule
A2J (Ours) & \textbf{8.57} & \textbf{6.92} & \textbf{9.95} & \textbf{105.06} \\
\bottomrule
\end{tabular}
\end{center}
\caption{Performance comparison on HANDS 2017 dataset~\cite{Hands2017}. ``SEEN" and ``UNSEEN" denote the cases whether the test subjects are involved in training set. ``AVG" indicates the result over all subjects. And, ``$*$" means the ensemble of 10 models.}
\label{tab:Hands2017}
\end{table}

\begin{table}
\scriptsize
\begin{center}
\begin{tabular}{ccc}
\toprule
Methods & Mean error (mm) & FPS\\
\hline\hline
DISCO \cite{DISCO} & 20.7 & - \\
Hand3D \cite{Hand3D} & 17.6 & 30 \\
DeepModel \cite{DeepModel} & 17.04 & - \\
JTSC \cite{JTSC} & 16.8 & - \\
Global-to-Local \cite{Global-to-Local} & 15.60 & 50 \\
Lie-X \cite{Lie-X} & 14.51 & - \\
REN-4x6x6 \cite{REN966} & 13.39 & - \\
REN-9x6x6 \cite{REN966} & 12.69 & - \\
DeepPrior++ \cite{Deepprior++} & 12.24 & 30 \\
Pose-REN \cite{Pose-REN} & 11.81 & - \\
HandPointNet \cite{HandPointNet} & 10.5 & 48 \\
{DenseReg} \cite{DenseReg} & 10.2 & 27.8 \\
V2V \cite{V2V} & 9.22 & 35 \\
P2P \cite{P2P} & 9.045 & 41.8 \\
\midrule
A2J (Ours) & \textbf{8.61} & \textbf{105.06} \\
\bottomrule
\end{tabular}
\end{center}
\caption{Performance comparison on NYU dataset~\cite{NYU}. ``Mean error" indicates the average 3D distance error.}
\label{tab:NYU}
\end{table}

\subsection{Comparison with state-of-the-art methods} \label{Experimente:2}

\textbf{HANDS 2017 dataset}: A2J is compared with the state-of-the-art 3D hand pose estimation methods~\cite{RCN3D,THUVCLab,HandPointNet,V2V} {, particularly}. The performance comparison is listed in Table~\ref{tab:Hands2017}. It can be observed that:

$\bullet$ On this challenging million-scale dataset, A2J consistently outperforms the other approaches both from the perspectives of effectiveness and efficiency. This essentially verifies the superiority of our proposition;

$\bullet$ It is worthy noting that, A2J is significantly superior to the others with the remarkable margin (2.05 at least) towards the ``UNSEEN" test case. This phenomenon essentially demonstrates the generalization {ability} of A2J;

$\bullet$ V2V$^*$ is the strongest competitor of A2J, but {with 10 models ensemble. As a consequence, it is much slower than A2J with only a single model.}


\begin{table}
\scriptsize
\begin{center}
\begin{tabular}{ccc}
\toprule
Methods & Mean error (mm) & FPS\\
\hline\hline
LRF \cite{ICVL} & 12.58 & - \\
DeepModel \cite{DeepModel} & 11.56 & - \\
Hand3D \cite{Hand3D} & 10.9 & 30 \\
CrossingNets \cite{CrossingNets} & 10.2 & 90.9 \\
Cascade \cite{Cascaded} & 9.9 & - \\
JTSC \cite{JTSC} & 9.16 & - \\
DeepPrior++ \cite{Deepprior++} & 8.1 & 30 \\
REN-4x6x6 \cite{REN966} & 7.63 & - \\
REN-9x6x6 \cite{REN966} & 7.31 & - \\
{DenseReg} \cite{DenseReg} & 7.3 & 27.8 \\
Pose-REN \cite{Pose-REN} & 6.79 & - \\
HandPointNet \cite{HandPointNet} & 6.935 & 48 \\
P2P \cite{P2P} & 6.328 & 41.8 \\
V2V$^*$ \cite{V2V} & \textbf{6.286} & 3.5 \\
\midrule
A2J (Ours) & 6.461 & \textbf{105.06} \\
\bottomrule
\end{tabular}
\end{center}
\caption{Performance comparison on ICVL dataset~\cite{ICVL}. ``Mean error" indicates the average 3D distance error.}
\label{tab:ICVL}
\end{table}

\begin{figure}
\centering
\subfigure {\label{sub_NYU} \includegraphics[height=3.1cm,width=8.1cm]{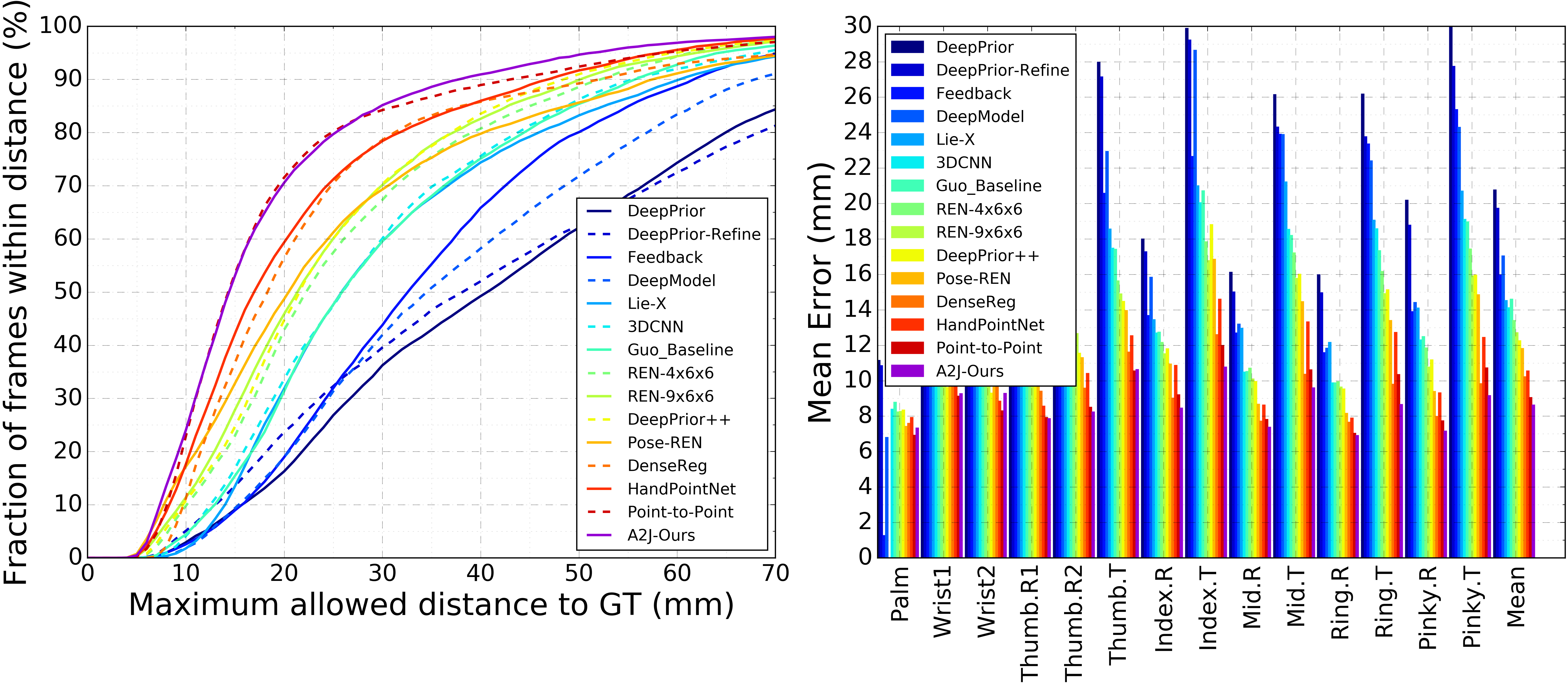}}
\subfigure {\label{sub_ICVL} \includegraphics[height=3.1cm,width=8.1cm]{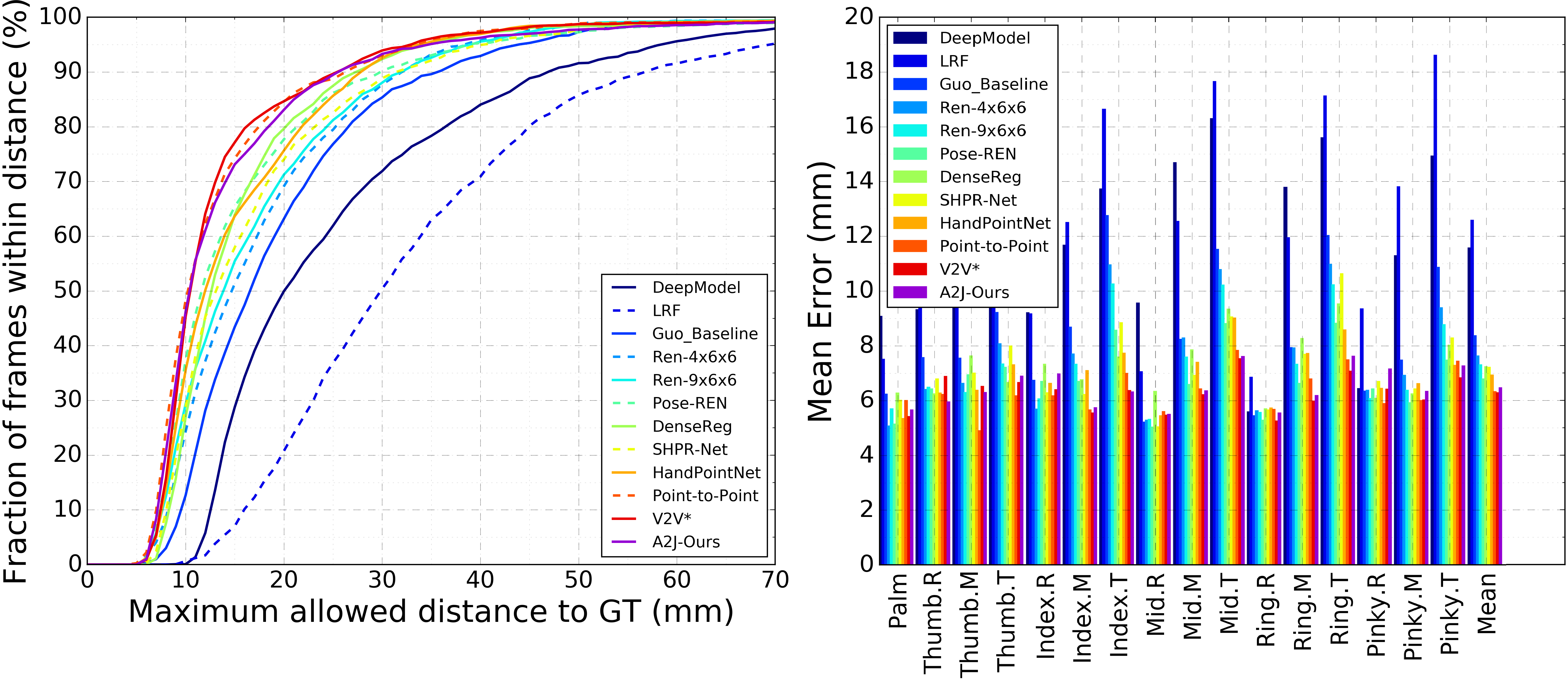}}
\caption{Comparison of A2J with state-of-the-art methods. Left: the percentage of success frames over different error thresholds. Right: 3D distance errors per hand keypoints. Top: NYU dataset. Bottom: ICVL dataset.}
\label{fig:rate_NYU_ICVL}
\end{figure}

\textbf{NYU and ICVL datasets}: We compare A2J with the state-of-the-art 3D hand pose estimation methods~\cite{ICVL,DISCO,Cascaded,Deepprior++,DeepModel,JTSC,Hand3D,Lie-X,REN466,REN966,CrossingNets,Pose-REN,Global-to-Local,DenseReg,HandPointNet,P2P,V2V} on this 2 datasets specifically. The experimental results are given in Table~\ref{tab:NYU}, ~\ref{tab:ICVL} on the average 3D distance error. Meanwhile, the percentage of success frames over different error thresholds and the error of each joint are also given in Fig.~\ref{fig:rate_NYU_ICVL}. We can summarize that:

$\bullet$ A2J is superior to the other methods in most cases both on accuracy and efficiency. The exceptional case is that, A2J is slightly inferior to V2V$^*$ and P2P on ICVL dataset on accuracy but with much higher running efficiency;

$\bullet$ Concerning the good tradeoff between effectiveness and efficiency, A2J essentially takes advantage over the state-of-the-art 3D hand pose estimation approaches.


\begin{table*}
\scriptsize
\begin{center}
\newcommand{\tabincell}[2]{
\begin{tabular}{@{}#1@{}}#2\end{tabular}
}
\centering
\setlength{\tabcolsep}{0.65mm}{
\begin{tabular}{c|cccccccc|ccccccc}
\toprule
 & \multicolumn{8}{c|}{mAP (front-view) } & \multicolumn{7}{c}{mAP (top-view) } \\
\hline\hline
Method     & \tabincell{c}{RF \\ \cite{RF}}   & \tabincell{c}{RTW \\ \cite{RTW}}  &
\tabincell{c}{ IEF\\ \cite{IEF}}  & \tabincell{c}{VI \\ \cite{VI}}   & \tabincell{c}{CMB \\ \cite{CMB}}  &
\tabincell{c}{REN-\\ 9x6x6 \cite{REN966}} & \tabincell{c}{V2V$^*$ \\ \cite{V2V}} & \tabincell{c}{A2J \\ (Ours)} &
\tabincell{c}{RF\\ \cite{RF}}   & \tabincell{c}{RTW \\ \cite{RTW}}  &
\tabincell{c}{IEF\\ \cite{IEF}}  & \tabincell{c}{VI \\ \cite{VI}} &
\tabincell{c}{REN-\\ 9x6x6 \cite{REN966}} & \tabincell{c}{V2V$^*$ \\ \cite{V2V}} & \tabincell{c}{A2J \\ (Ours)} \\
\hline\hline
Head       & 63.8 & 97.8 & 96.2 & 98.1 & 97.7 & {98.7} & 98.29 & 98.54   & 95.4 & {98.4} & 83.8 & 98.1 & 98.2 & {98.4} & 98.38 \\
Neck       & 86.4 & 95.8 & 85.2 & 97.5 & 98.5 & {99.4} & 99.07 & 99.20   & 98.5 & 82.2 & 50.0 & 97.6 & 98.9 & {98.91} & {98.91} \\
Shoulders  & 83.3 & 94.1 & 77.2 & 96.5 & 75.9 & 96.1 & {97.18} & 96.23   & 89.0 & 91.8 & 67.3 & 96.1 & 96.6 & {96.87} & 96.26 \\
Elbows     & 73.2 & 77.9 & 45.4 & 73.3 & 62.7 & 74.7 & {80.42} & 78.92   & 57.4 & 80.1 & 40.2 & {86.2} & 74.4 & 79.16 & 75.88 \\
Hands      & 51.3 & {70.5} & 30.9 & 68.7 & 84.4 & 55.2 & 67.26 & 68.35   & 49.1 & 76.9 & 39.0 & {85.5} & 50.7 & 62.44 & 59.35 \\
Torso      & 65.0 & 93.8 & 84.7 & 85.6 & 96.0 & 98.7 & {98.73} & 98.52   & 80.5 & 68.2 & 30.5 & 72.9 & {98.1} & 97.78 & 97.82 \\
Hips       & 50.8 & 90.3 & 83.5 & 72.0 & 87.9 & 91.8 & {93.23} & 90.85   & 20.0 & 55.7 & 38.9 & 61.2 & 85.5 & {86.91} & 86.88 \\
Knees      & 65.7 & 68.8 & 81.8 & 69.0 & 84.4 & 89.0 & {91.80} & 90.75   & 2.6 & 53.9 & 54.0 & 51.6 & 70.0 & {83.28} & 79.66 \\
Feet       & 61.3 & 68.4 & 80.9 & 60.8 & 83.8 & 81.1 & {87.60} & 86.91   & 0.0 & 28.7 & 62.4 & 51.5 & 41.6 & {69.62} & 58.34 \\
\hline\hline
mean  & 65.8 & 80.5 & 71.0 & 77.4 & 83.3 & 84.9 & \textbf{88.74} & 88.0 & 47.4 & 68.2 & 51.2 & 75.5 & 75.5 & \textbf{83.44} & 80.5 \\
\bottomrule
\end{tabular}}
\end{center}
\caption{Performance comparison on ITOP 3D body pose estimation dataset~\cite{VI}.}
\label{tab:ITOP}
\end{table*}

\textbf{ITOP dataset}: We also compare A2J with the state-of-the-art 3D body pose estimation manners~\cite{RF,RTW,IEF,VI,REN966,CMB,V2V} on this dataset. The performance comparison is listed in Table~\ref{tab:ITOP}. We can see that:

$\bullet$ A2J is significantly superior to the other ones both for front-view and top-view tracks, except V2V$^*$. The performance gap is 3.1 at least for front-view case, and 5 at least for top-view case. This reveals that A2J is also applicable to 3D body pose estimation, as well as 3D hand task;

$\bullet$ A2J is inferior to V2V$^*$. However, V2V$^*$ actually consists of 10 models ensemble . Thus, compared with A2J with single model it is of much lower running efficiency.


\textbf{K2HPD dataset}: Since this body pose dataset only provides the pixel-level in-plain ground-truth, the depth estimation branch within A2J is removed accordingly. We also compare A2J with the state-of-the-art approaches~\cite{PHR,CPM,StackedHG,K2HPD,CMB}. The performance comparison is given in Table~\ref{tab:K2HPD}. It can be observed that:

$\bullet$ A2J outperforms the other methods by large margins consistently, corresponding to the difference PDJ thresholds. In average, the performance gap is 10.8 at least. This demonstrates that, A2J is also applicable to 2D case;

$\bullet$ It is worthy noting that, with the decrease of PDJ threshold the advantage of A2J will be enlarged remarkably. This reveals the fact that, A2J is essentially superior to more accurate body pose estimation.

\begin{table}
\scriptsize
\begin{center}
\newcommand{\tabincell}[2]{
\begin{tabular}{@{}#1@{}}#2\end{tabular}
}
\centering
\setlength{\tabcolsep}{1.0mm}{
\begin{tabular}{ccccccc}
\toprule
Method & \tabincell{c}{PHR \\ \cite{PHR}} & \tabincell{c}{CPM \\ \cite{CPM}} & \tabincell{c}{SH \\ \cite{StackedHG}} &
\tabincell{c}{IEML \\ \cite{K2HPD}} & \tabincell{c}{CMB \\ \cite{CMB}} & \tabincell{c}{A2J \\ (Ours)} \\
\hline \hline
PDJ (0.05) & 26.8 & 30.0 & 41.0 & 43.2 & 52.5  & \textbf{76.3} \\
PDJ (0.10) & 70.3 & 58.5 & 73.7 & 64.1 & 84.2  & \textbf{94.4} \\
PDJ (0.15) & 84.7 & 87.8 & 84.6 & 88.1 & 91.7  & \textbf{97.6} \\
PDJ (0.20) & 91.3 & 93.6 & 89.0 & 91.0 & 95.1 & \textbf{98.6} \\
\hline
Average & 68.3 & 67.5 & 72.1 & 71.6 & 80.9 & \textbf{91.7} \\
\bottomrule
\end{tabular}}
\end{center}
\caption{Performance comparison on K2HPD dataset~\cite{K2HPD}.}
\label{tab:K2HPD}
\end{table}

\subsection{Ablation study}
\label{Experimente:3}
\subsubsection{Component effectiveness analysis} \label{ablation:component}

The component effectiveness analysis within A2J is executed on NYU~\cite{NYU} (hand), and ITOP~\cite{VI} dataset (body). We will investigate the effectiveness of anchor proposal branch, informative anchor point surrounding loss, and configuration of in-plain offset and depth estimation branches. The results are listed in Table~\ref{tab:loss12}. It can be observed that:

$\bullet$ Without using anchor proposal branch, performance will drop remarkably especially for body pose. This verifies our point that, not all the anchor points contribute equally to the certain joints. Actually, anchor point adaptivity is A2J's essential property to leverage performance;

$\bullet$ Without using informative anchor point surrounding loss, performance will drop especially for body pose. This demonstrate that, informative anchor point spatial layout is an essential issue that should be concerned towards generalization ability;

$\bullet$ When estimating in-plain offset and depth value in one branch, performance will drop to some degree. This may be caused by the fact that, in-plain offset and depth value holds different physical characteristics.

\begin{table}[]
\scriptsize
\begin{center}
\begin{tabular}{ccc}
\toprule
Dataset                                                                    & Component                      & error / mAP            \\ \hline \midrule
\multirow{4}{*}{\begin{tabular}[c]{@{}c@{}}NYU \\ (hand)\end{tabular}}    & \multicolumn{1}{c}{w/o anchor proposal branch}  & \multicolumn{1}{c}{10.08} \\ \cmidrule{2-3}
                                                                           & w/o informative anchor point surrounding loss              &    9.00              \\ \cmidrule{2-3}
                                                                           & Estimate IPO and DV using one branch         & 8.95               \\ \cmidrule{2-3}
                                                                           & \multicolumn{1}{c}{A2J (Ours)} & \multicolumn{1}{c}{\textbf{8.61}} \\ \hline \midrule
\multirow{4}{*}{\begin{tabular}[c]{@{}c@{}}ITOP \\ front-view \\ (body pose)\end{tabular}} &   w/o anchor proposal branch                 & 80.1                 \\ \cmidrule{2-3}
                                                                           & \multicolumn{1}{c}{w/o informative anchor point surrounding loss}  & \multicolumn{1}{c}{86.4} \\  \cmidrule{2-3}
                                                                           & \multicolumn{1}{c}{Estimate IPO and DV using one branch}  & \multicolumn{1}{c}{87.4} \\ \cmidrule{2-3}
                                                                           & A2J (Ours)                     & \textbf{88.0}               \\ \bottomrule
\end{tabular}
\end{center}
\caption{Component effectiveness analysis within A2J. ``IPO" indicates in-plain offset, and ``DV" denotes depth value.}
\label{tab:loss12}
\end{table}

\subsubsection{Effectiveness of anchor-based paradigm} \label{ablation:1}

To verify the effectiveness of anchor-based 3D pose estimation paradigm, we compare A2J with the global regression based manner~\cite{Deeppose} and FCN-based approach~\cite{MSbaseline}. Since FCN model is generally used to predict in-plain joint position, this ablation study is executed on K2HPD dataset~\cite{K2HPD} only with in-plain ground-truth annotation. Global regression manner encodes depth image with 2D CNN, and then regresses in-plain human joint position using fully-connected layers. FCN model is built following~\cite{MSbaseline}. ResNet-50~\cite{ResNet} is employed as the backbone network for them, which is the same as A2J for fair comparison. PDJ (0.05) is used as the evaluation criteria. The performance comparison is listed in Table~\ref{tab:FCN_Global}. We can see that:

$\bullet$ Our proposed anchor-based paradigm significantly outperforms the other 2 ones, when using the same ResNet-50 backbone network. We think 2 main reasons lie. First, compared with global regression based manner local spatial context information can be better maintained within A2J. Meanwhile, compared with FCN model A2J possess anchor point adaptivity towards the certain joint;

$\bullet$ A2J runs faster than FCN model, but slower than global regression way. However, its performance advantage over global regression paradigm is significant, actually with better tradeoff between effectiveness and efficiency.

\begin{table}
\scriptsize
\begin{center}
\begin{tabular}{cccc}
\toprule
Paradigm & Global regression~\cite{Deeppose} & FCN model~\cite{MSbaseline} & A2J (Ours) \\
\hline\hline
PDJ (0.05) & 61.5 & 70.4 & \textbf{76.3} \\
\midrule
FPS & \textbf{145.12} & 45.48 & 93.78 \\
\bottomrule
\end{tabular}
\end{center}
\caption{Performance comparison among the different paradigms on K2HPD dataset~\cite{K2HPD}.}
\label{tab:FCN_Global}
\end{table}

\subsubsection{Effectiveness of the pre-training} \label{ablation:2}

One reason for why we build A2J on 2D CNN is that, it can be pre-trained on the large-scale RGB visual datasets (e.g., ImageNet) for knowledge transfer. To verify this point, we compare the performance of A2J with and without pre-training on ImageNet on NYU (hand) and ITOP (body) datasets. The performance comparison is listed in Table~\ref{tab:pretrain}. It can be observed that, both for hand and body pose cases pre-training A2J on ImageNet can indeed help to leverage the performance.

\begin{table}
\scriptsize
\begin{center}
\begin{tabular}{ccc}
\toprule
Pre-train & From scratch & ImageNet pre-training \\
\hline\hline
NYU (error) & 10.08 & \textbf{8.61} \\
\midrule
ITOP front-view (mAP) & 87.3 & \textbf{88.0} \\
\bottomrule
\end{tabular}
\end{center}
\caption{Effectiveness of pre-training A2J on ImageNet.}
\label{tab:pretrain}
\end{table}

\begin{table}[]
\scriptsize
\begin{center}
\begin{tabular}{ccccc}                                                     \toprule
\multicolumn{1}{c}{}                                                       & Backbone & ResNet-18 & ResNet-34 & ResNet-50 \\  \hline\midrule
\multirow{2}{*}{NYU}                                                       & error    & 9.32      & 9.01      & \textbf{8.61}      \\  \cmidrule {2-5}  
                                                                           & FPS      & \textbf{192.25}    & 144.63    & 105.06    \\ \hline \midrule
\multirow{2}{*}{\begin{tabular}[c]{@{}c@{}}ITOP\\ front-view\end{tabular}} & mAP      & 87.1      & 87.8      & \textbf{88.0}      \\   \cmidrule {2-5}
                                                                           & FPS      & \textbf{167.19}    & 122.47    & 93.78   \\ \bottomrule
\end{tabular}
\end{center}
\caption{Performance comparison among the backbones.}
\label{tab:backbone}
\end{table}

\subsubsection{Backbone network comparison} \label{ablation:3}

The comparison among the different backbone networks is further studied. As shown in Table~\ref{tab:backbone}, we compare the performance of 3 backbone networks (i.e., ResNet-18, ResNet-34 and ResNet-50). It can be summarized that:

$\bullet$ Deeper network can achieve better results, but with relatively slower running efficiency. However, the performance gap among the different backbones is not huge;

$\bullet$ It is worthy noting that, even using ResNet-18 A2J still can {generally} achieve the state-of-the-art performance and with extremely fast running speed of 192.25 FPS. This reveals the applicability of A2J towards high real-time running demanding application scenarios.


\subsection{Qualitative evaluation} \label{Qualitative}

\begin{figure}
\centering
\subfigure [Qualitative results on NYU dataset] {\label{sub_NYU} \includegraphics[width=6.3cm]{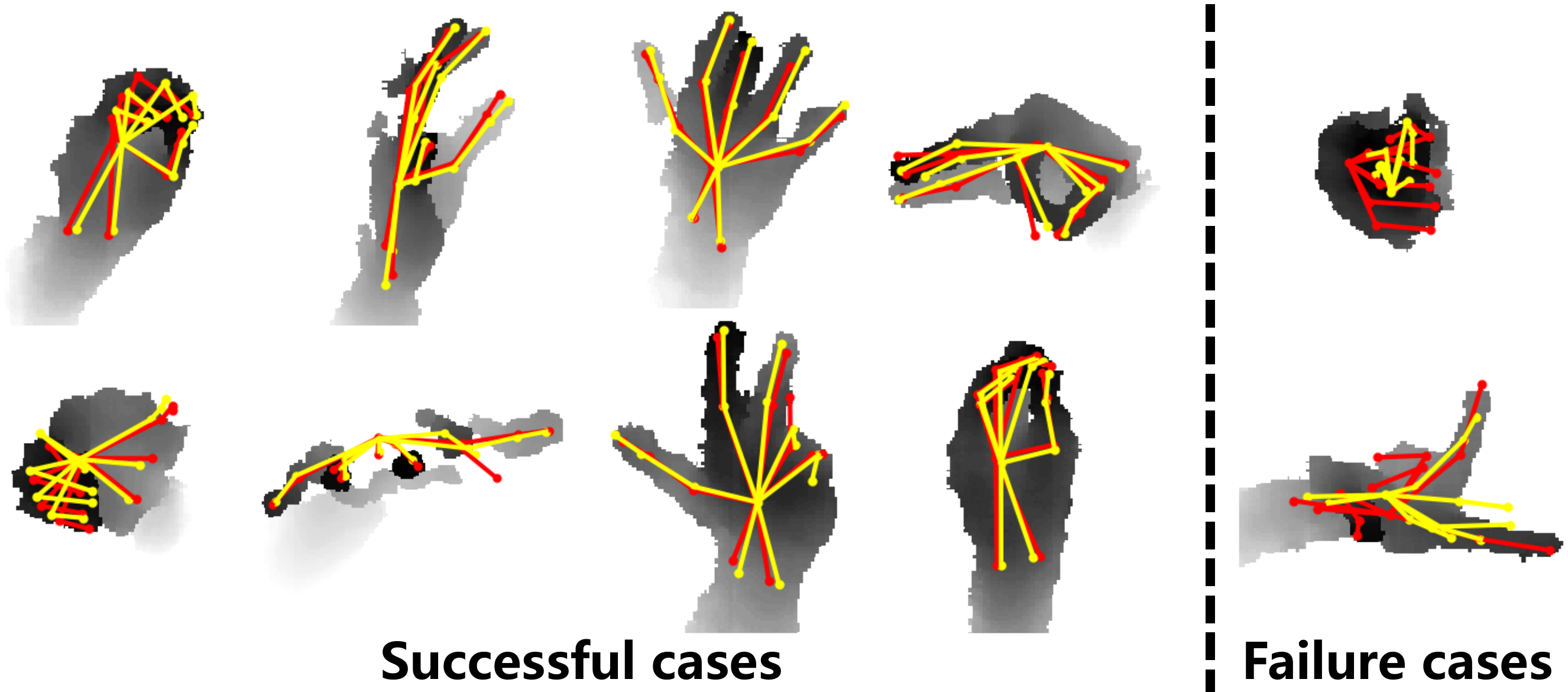}}
\subfigure [Qualitative results on ITOP front-view dataset] {\label{sub_ICVL} \includegraphics[width=6.3cm]{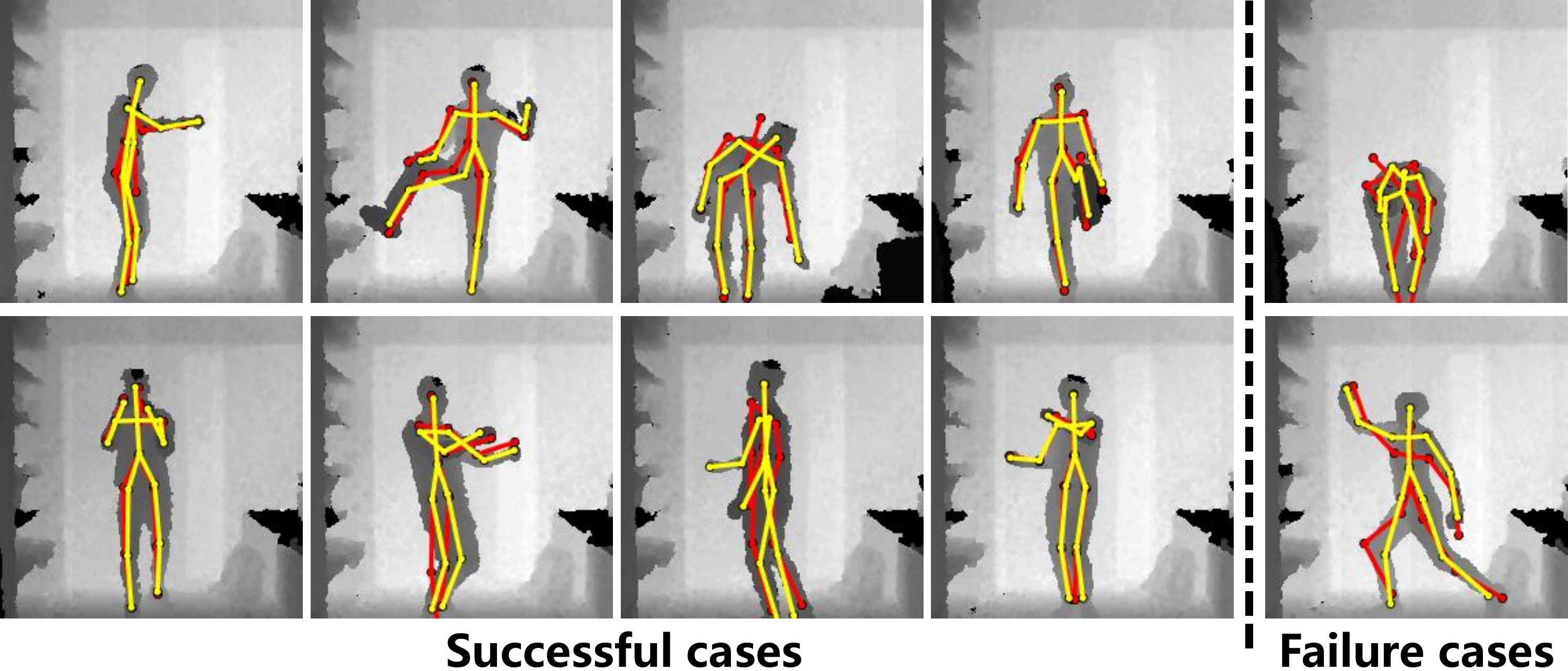}}
\caption{Qualitative results of A2J. Ground-truth is shown in red, and the predicted pose is in yellow.}
\label{fig:Qualitative_nyu_itop}
\end{figure}

Some qualitative results of A2J on NYU~\cite{NYU} and ITOP (front-view)~\cite{VI} datasets are shown in Fig.~\ref{fig:Qualitative_nyu_itop}. We can see that, generally A2J works well both for 3D hand and body pose estimation. The failure cases are mainly caused by the serious self-occlusion and dramatic pose variation.

\subsection{Running speed analysis} \label{Experimente:4}

The average online running speed of A2J for 3D hand pose estimation is 105.06 FPS, including 1.5 ms for reading and warpping image, and 8.0 ms for network forward propagation and post-processing on a single NVIDIA 1080Ti GPU. The running speed for 3D body pose estimation is 93.78 FPS, including 0.4 ms for reading and warpping image, and 10.2 ms for network forward propagation and post-processing. This reveals A2J's real-time running capacity.

\section{Conclusions} \label{Conclusion}

In this paper, an anchor-based 3D articulated pose estimation approach for single depth image termed A2J is proposed. Within A2J anchor points are densely set up on depth image to capture the global-local spatial context information, and predict joint's position in ensemble way. Meanwhile, informative anchor points are extracted to reveal A2J's adaptive characteristics towards the different joints. A2J is built on 2D CNN without using computational expensive deconvolutional layers. The wide-range experiments demonstrate A2J's superiority both from the perspectives of effectiveness and efficiency. In future work, we will seek the more effective way to fuse the anchor points.

\section*{Acknowledgment}
This work is jointly supported by the National Key R\&D Program of China (No. 2018YFB1004600), National Natural Science Foundation of China (Grant No. 61876211 and 61602193), the Fundamental Research Funds for the Central Universities (Grant No. 2019kfyXKJC024), the International Science \& Technology Cooperation Program of Hubei Province, China (Grant No. 2017AHB051), {the start-up funds from University at Buffalo.} Joey Tianyi Zhou is supported by Singapore Government's Research, Innovation and Enterprise 2020 Plan (Advanced Manufacturing and Engineering domain) under Grant A1687b0033 and Grant A18A1b0045. We also thank the anonymous reviewers for their suggestions to enhance the quality of this paper.

\newpage

{\small
\bibliographystyle{ieee_fullname}
\bibliography{egbib}
}

\end{document}